\documentclass[journal]{IEEEtran}
\usepackage{amsmath,graphicx}
\usepackage{booktabs}
\usepackage{cite}
\usepackage{multirow}
\usepackage{multicol}
\usepackage{tabularx}
\usepackage{float}
\usepackage{makecell}
\usepackage{subfigure}
\usepackage{enumitem}
\usepackage{amsfonts,amssymb}
\usepackage[colorlinks]{hyperref}
\usepackage{cleveref}
\usepackage{bm}

\crefname{figure}{Fig.}{Figs.}
\Crefname{figure}{Figure}{Figures}
\crefname{section}{Sect.}{Sects.}
\Crefname{section}{Section}{Section}
\crefname{equation}{Eq.}{Eqs.}
\Crefname{equation}{Equation}{Equation}
\crefname{table}{Table}{Tables.}
\Crefname{table}{Table}{Tables}

\newcommand{\etal}{\textit{et al}.}

\ifCLASSINFOpdf
\else
\fi
\begin{document}
%
\title{Unsupervised Cycle-consistent Generative Adversarial Networks for Pan-sharpening}
%
%
%

\author{Huanyu~Zhou,
		Qingjie~Liu,~\IEEEmembership{Member,~IEEE,}
		Dawei~Weng,
        and~Yunhong~Wang,~\IEEEmembership{Fellow,~IEEE}
}

%



\maketitle

\begin{abstract}
Deep learning based pan-sharpening has received significant research interest in recent years. Most of existing methods fall into the supervised learning framework in which they down-sample the multi-spectral (MS) and panchromatic (PAN) images and regard the original MS images as ground truths to form training samples. Although impressive performance could be achieved, they have difficulties generalizing to the original full-scale images due to the scale gap, which makes them lack of practicability. In this paper, we propose an unsupervised generative adversarial framework that learns from the full-scale images without the ground truths to alleviate this problem. We extract the modality-specific features from the PAN and MS images with a two-stream generator, perform fusion in the feature domain, and then reconstruct the pan-sharpened images. Furthermore, we introduce a novel hybrid loss based on the cycle-consistency and adversarial scheme to improve the performance. Comparison experiments with the state-of-the-art methods are conducted on GaoFen-2 and WorldView-3 satellites. Results demonstrate that the proposed method can greatly improve the pan-sharpening performance on the full-scale images, which clearly show its practical value. Codes are available at https://github.com/zhysora/UCGAN.
\end{abstract}


%
\IEEEpeerreviewmaketitle

\section{Introduction}

\begin{figure*}[t]
\centering
	\subfigure{
		\includegraphics[width=.32\linewidth]{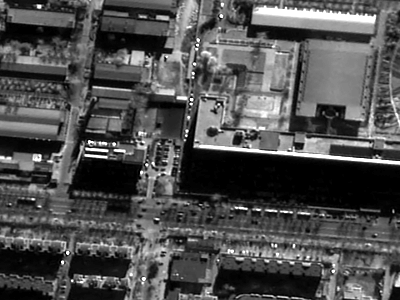}} \hspace{-9pt}
	\subfigure{
		\includegraphics[width=.32\linewidth]{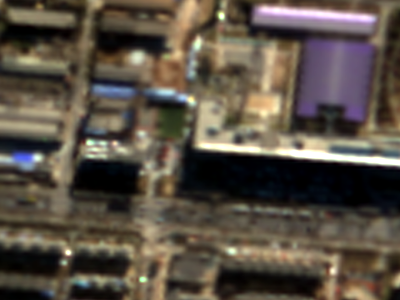}} \hspace{-9pt}
	\subfigure{
		\includegraphics[width=.32\linewidth]{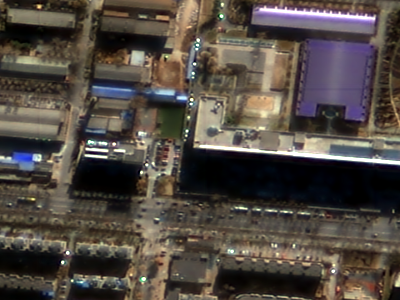}} \\
		\vspace{-8pt}
	\addtocounter{subfigure}{-3}
	\subfigure[]{
		\includegraphics[width=.32\linewidth]{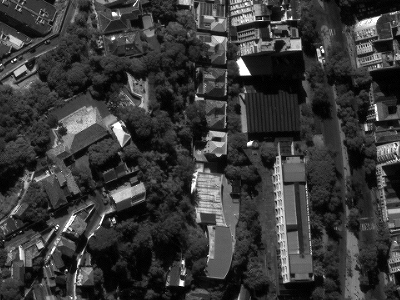}} \hspace{-9pt}
	\subfigure[]{
		\includegraphics[width=.32\linewidth]{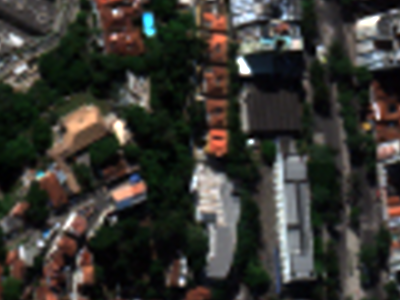}} \hspace{-9pt}
	\subfigure[]{
		\includegraphics[width=.32\linewidth]{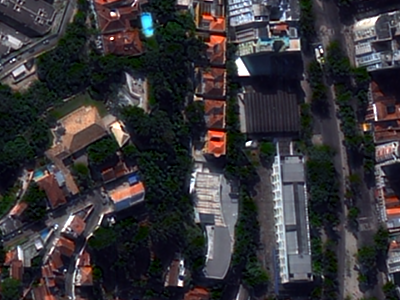}} \\
		\vspace{-8pt}
	\caption{Examples of the pan-sharpening results. (c) is the pan-sharpened result from (a) PAN and (b) LR MS. Images are selected from our GaoFen-2 and WorldView-3 datasets. Visualized in RGB.}
\label{fig::example}
\end{figure*}

\IEEEPARstart{M}ulti-resolution images are quite common in remote sensing systems as they offer users a pair of modalities: multi-spectral (MS) images and panchromatic (PAN) images. These two modalities are at the highest resolution either in the spectral or in the spatial domain. However, due to physical constraints\cite{physicalConstraint}, these goals cannot be achieved at the same time. As a result, the MS images are at lower spatial resolution and the PAN images are with poor spectral information. To meet the demands for high-resolution multi-spectral (HR MS) images in a number of practical applications, the pan-sharpening task, whose aim is to combine the strengths of both the MS and PAN images to generate the desired HR MS images, has been receiving great attention from the community and entered a booming period. Fig.~\ref{fig::example} displays some examples of our pan-sharpening results.

In recent years, deep learning based methods have obtained impressive achievements in a large number of computer vision tasks, motivating the pan-sharpening community to leverage the power of the deep neural network architectures. The first attempt is PNN \cite{PNN}, who draw inspirations from SRCNN \cite{SRCNN} and modify it by embedding domain-specific knowledge of remote sensing field. The follow up works embrace the advances in deep learning and achieve remarkable progress. For instance, inspired by residual learning \cite{ResNet}, PanNet \cite{PanNet} proposes a much deeper network than PNN. It employs the skip-connection idea of ResNet \cite{ResNet} to design deeper networks and propose to make full use of the high-frequency information to produce clear spatial details. Sharing a similar spirit, DRPNN \cite{DRPNN} also designs a deeper network to strengthen the model capacity. In addition to deepen the neural network, MSDCNN \cite{MSDCNN} explores different sizes of CNN filters and proposes a multi-scale architecture. Considering PAN and MS images contain distinct information, TFNet \cite{TFNet} proposes to extract features of them using two stream networks and performs fusion in the feature domain. Viewing pan-sharpening as an image generation problem, PSGAN \cite{PSGAN} takes advantage of  generative adversarial learning \cite{GAN} and improves the fusion quality, especially visual quality of TFNet. Thanks to the powerful deep learning techniques and large amounts of remote sensing data, pan-sharpening has been advancing rapidly.

Nevertheless, existing deep learning based pan-sharpening methods still face great challenges when applying them to real applications. One limitation is that they require ideal HR MS images, which do not exist, to optimize the networks. Not the best, but doable, they follow the Wald's protocol \cite{wald} to train their models, i.e., both the PAN and MS images are down-sampled to a lower scale so that the original MS images serve as the ground truth images. As a result, they should generalize to full-scale images in practical applications. These methods provide satisfactory results on the down-scaled images with a minimal effort, however, face performance degradation on the original full-scale images, which makes the supervised learning methods impractical. To be specific, there is a gap between the down-scaled and full-scale remote sensing images. Different from natural images, remote sensing images usually contain much deeper bit depth and are with very different characteristics. Our motivation is to make full use of the original information of the remote sensing images without any preprocessing since the downsampling process may alter the original characteristics of images. To handle this problem, we propose an unsupervised generative adversarial network for pan-sharpening, termed UCGAN. UCGAN focuses on unsupervised learning and is trained on the original full-scale images to perform pan-sharpening. To better extract features from inputs and learn the mapping from inputs to the fused images, we learn from the ascendancy of the existing work  \cite{PSGAN}. Firstly, we extract the modality-specific features from the PAN and MS images with a two-stream generator and conduct the fusion in the feature domain. To better preserve spatial information, we extract the high-frequency information from images and then feed them into the two-stream generator. Finally, we introduce a novel hybrid loss to boost the quality of the pan-sharpened images. The loss function is based on the cycle-consistency between the fusion output and the PAN and MS inputs. It should be noted that our network is trained without ground truth images. Our major contributions can be summarized as follows: 
\begin{itemize}
\item We design an unsupervised cycle-consistent generative adversarial network for pan-sharpening, termed UCGAN, which can be trained on the full-scale images to take advantage of the rich spatial and spectral information of the original PAN and MS images. 
\item We introduce a novel hybrid loss based on cycle-consistency for optimizing the network under the unsupervised learning framework. No ground truth images are required during training. 
\item Extensive experiments on GaoFen-2 and WorldView-3 datasets demonstrate that our proposed method greatly improves the performance on the original full-scale images and provides the best results with clear spatial and spectral details.

\end{itemize}  

\begin{figure*}[t]
  \centering
  \includegraphics[width=\linewidth]{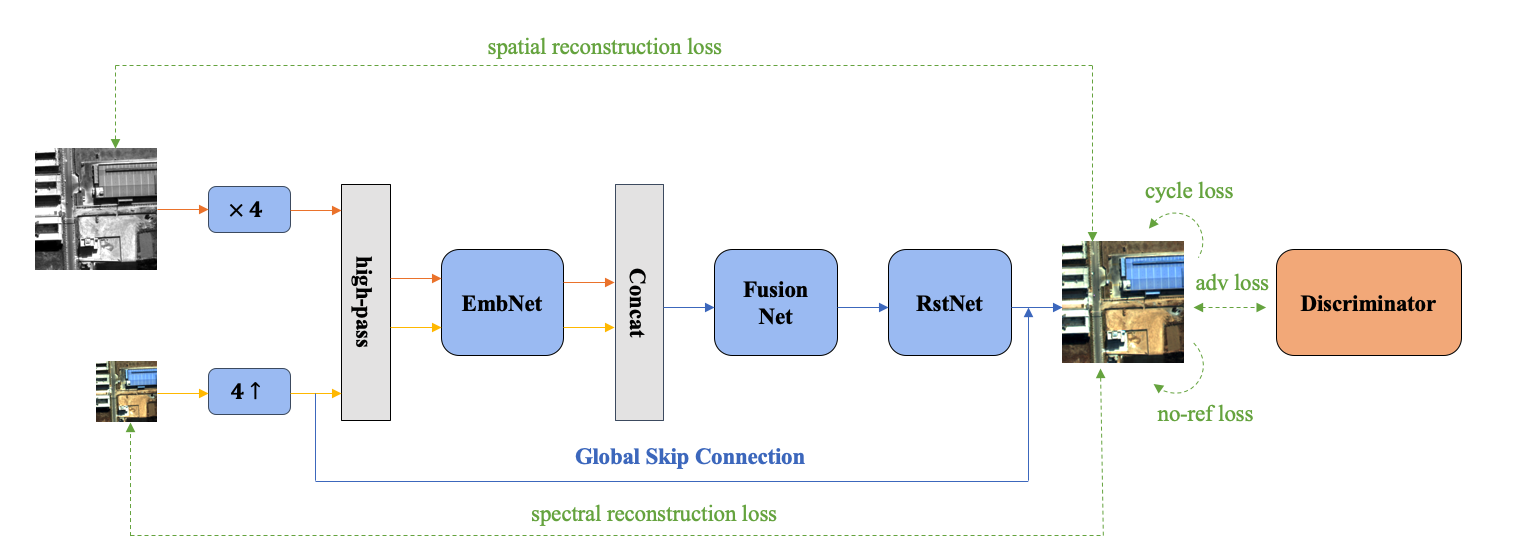}
  \caption{An overview of our unsupervised framework for pan-sharpening.``$\times 4$" means replicating the image 4 times in the channel dimension. ``$4 \uparrow$" stands for up-sampling with the ratio of 4 in the spatial resolution. ``high pass" is a filter to extract the high-frequency information from images. The solid lines present the flows of the data while the dotted lines stand for the loss calculations.}
  \label{fig::overview}
\end{figure*} 

\section{Related Work} \label{section::Related Work}
\subsection{Deep Learning based Pan-sharpening}
Inspired by the achievements of deep learning techniques in various computer vision tasks, especially single-image super-resolution, more and more researchers try to use deep learning models to solve the pan-sharpening problem. Masi \etal \cite{PNN} pioneer this field with a three-layer CNN whose main architecture is borrowed from a super-resolution model SRCNN\cite{SRCNN}. Following the spirit of PNN \cite{PNN}, Zhong \etal \cite{hybridPNN} improve it with a two-stage strategy: they first super-resolving the LR MS image with a CNN method and then fuse the spatial-enhanced MS and PAN using Gram–Schmidt transform to obtain the pan-sharpened image. Recently, deeper neural networks have been suggested to produce a better performance on vision tasks \cite{deeper1, deeper2}. Refer to ResNet \cite{ResNet}, Yang \etal \cite{PanNet} design a much deeper network with the help of residual learning. Similar ideas are adopted in \cite{residualcnnforpansharp1} and \cite{residualcnnforpansharp2}, they achieve good preservation of spatial and spectral information from images. From a different perspective, Deng \etal \cite{DiCNN} explore the combination of deep learning techniques and tradition fusion schemes, such as component substitution and multi-resolution analysis, and propose a detail injection-based network. For better information preservation, Han \etal \cite{SDPNet} use two encoder-decoder networks to learn enhanced representations from PAN and MS images, respectively. Xie \etal \cite{HPGAN} design 3D CNN architectures and reduce the sensitivity to different sensors by training in the high-frequency domain. 

\subsection{Generative Adversarial Networks}
Generative adversarial networks (GANs) are generative models introduced by Goodfellow \etal \cite{GAN}. GANs have been applied to many image generation-related tasks and achieve excellent performances \cite{MsCGAN, SAGAN}. A GAN is consist of two components, a generator $G$ and a discriminator $D$. $G$ focuses on generating images that are close to the real data, while $D$ aims to distinguish them from real ones. The main idea of it is training $G$ and $D$ adversarially so that $G$ is able to generate more and more realistic images and $D$ becomes stronger at the same time. After \cite{GAN}, many variations have been developed to improve the performance of GANs or perfect them by solving problems such as model collapse. For instance, Radford \etal \cite{DCGAN} propose DCGAN by replacing MLP with CNN, which strengths feature representative ability of GANs. Mao \etal \cite{LSGAN} make a modification to the original loss function and introduce a new one based on least-squares, which is beneficial to solve the vanishing gradients problem. Likewise, Arjovsky \etal \cite{WGAN} redefine the loss function based on the Wasserstein distance and weight clipping, which is shown to be helpful to stabilize the training process of GANs. Gulrajani \etal \cite{WGAN-GP} further overcome the instability training problem by punishing norms of the gradients of the discriminator w.r.t. its input. In pan-sharpening task, researchers also adopt GANs to improve the fusion performance. Liu \etal \cite{PSGAN} are one of the pioneers introducing GANs to the pan-sharpening task. Their model could produce high quality pan-sharpened images. Ma \etal \cite{Pan-GAN} handle pan-sharpening as a multi-task problem and train an unsupervised model with dual-discriminator for spatial and spectral preservation to generate pan-sharpened images with less spatial and spectral distortions.

\subsection{Unsupervised Pan-sharpening methods}
Recently, pan-sharpening methods based on unsupervised learning are of great concern, because they can be trained without ground truth and is of great value to the practical application. Luo \etal \cite{luo2020pansharpening} develop a novel unsupervised CNN-based pan-sharpening method without using any hand-crafted labels. Qu \etal \cite{UP-SAM} propose an unsupervised pan-sharpening method based on the self-attention mechanism to estimate the spatial varying detail extraction and injection functions. Zhou \etal \cite{PercepPan} introduce perceptual loss \cite{PercepLoss} to pan-sharpening and suggest a novel training paradigm that combines both supervised and unsupervised techniques. They first pre-train their model in a supervised manner and then fine-tune it in an unsupervised manner. Ma \etal \cite{Pan-GAN} propose a novel unsupervised framework for pan-sharpening based on a generative adversarial network to preserve rich spectral and spatial information from inputs.

\section{Method} \label{section::Method}

\begin{figure*}[t]
  \centering
  \includegraphics[width=.85\linewidth]{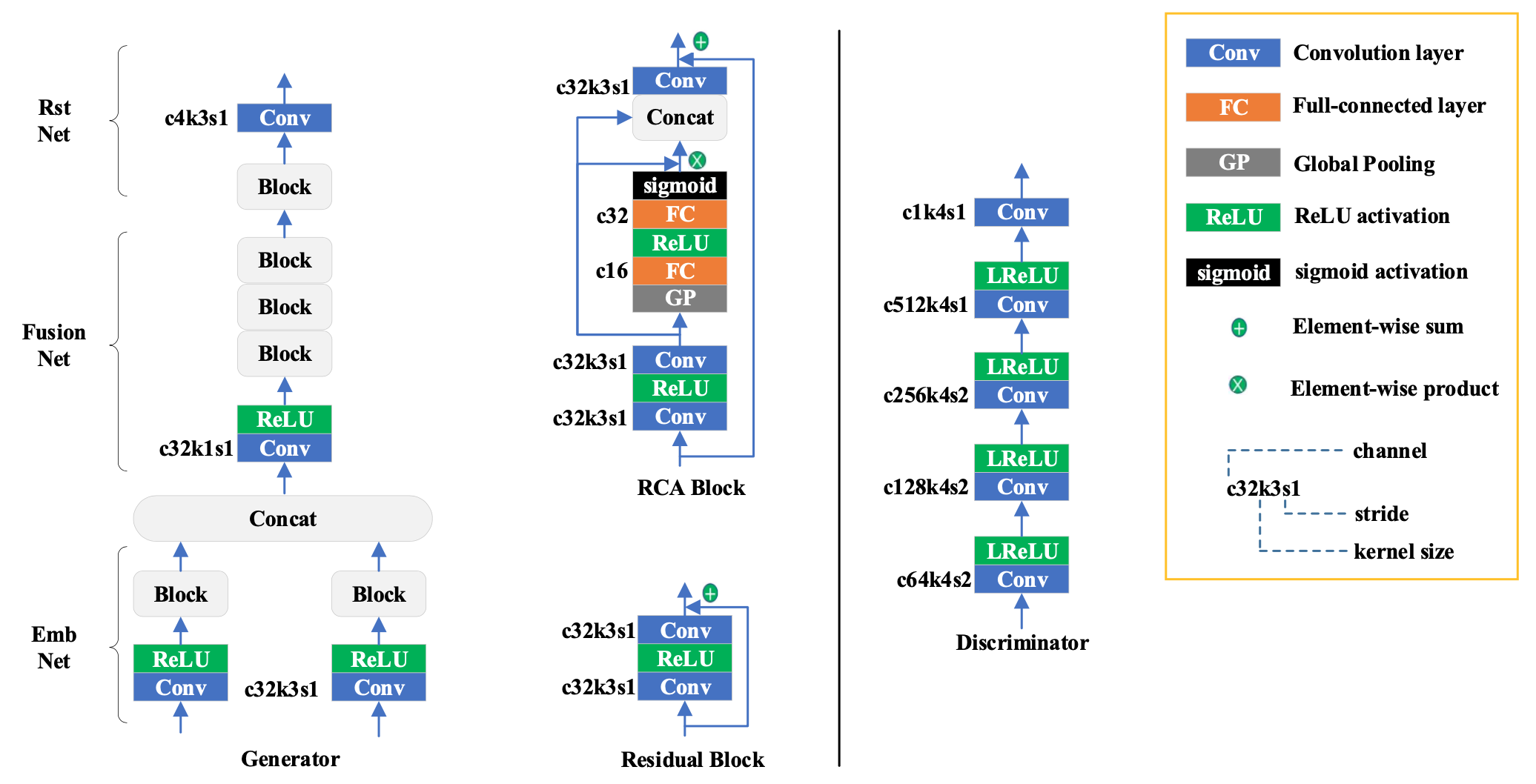}
  \caption{Detailed architectures of the Generator G (on the left) and Discriminator D (on the right). The parameters of each layer are given on the side, where ``c32k3s1" defines a $3 \times 3$ convolution layer with a stride of 1 whose output channel is 32. Our generator can be divided into 3 parts, EmbNet, FusionNet, and RstNet. We have tried two different modules, Residual Block \cite{ResNet} and RCA Block \cite{RCAGAN} in our generator and the experiments show RCA Block can improve the performance.}
  \label{fig::architecture}
\end{figure*}

\subsection{Overview}
Fig.~\ref{fig::overview} shows an overview of our unsupervised framework for pan-sharpening. Firstly, due to the distinct shapes of the input pair, we make an alignment between them. The PAN image replicates 4 times along the channel dimension and the LR MS image is up-sampled to the same spatial resolution as the PAN image using bicubic interpolation. As a result, both two types of images share the same shape. The alignment is necessary for using a uniform neural network to extract features and accomplish fusion in the subsequent procedure. After that, a high-pass filter works on the images to extract the high-frequency information for better spatial information preservation. Similar to \cite{PanNet}, we subtract the low-frequency content acquired by an averaging filter from the original images to obtain this high-frequency information. The processed PAN and MS images are then mapped into the feature domain by the embedding network, called EmbNet, who outputs the corresponding embedding vectors in a higher dimension. It is noted that before EmbNet, there are no trainable parameters and all operations are differentiable. After that, the fusion network, termed as FusionNet, takes the concatenation of the two embeddings as input and fuses them to one fused embedding vector. Finally, the restoration network, termed as RstNet, transfers the embedding vector back to the image domain to produce the pan-sharpened image. Additionally, to make the network stable and easy to train, there is a global skip connection from the interpolated LR MS to the final output. It has been discussed in \cite{PanNet} and we keep this manner. As an unsupervised method, there are no additional supervisions in our framework. To improve the quality of our fusion results, we introduce cycle-consistency and adversarial learning to our framework and add some restrictions to force the pan-sharpened result to keep the spatial and spectral information. As a result, we design a novel hybrid loss function which mainly consists of four parts, including cycle loss, adversarial loss, reconstruction loss, and non-reference loss. The details of them will be discussed in the following sections.

\subsection{Architecture}
Fig.~\ref{fig::architecture} displays the detailed architecture of our neural networks, which can be divided into two parts: generator and discriminator. 

\subsubsection{Generator}
All convolution layers in the generator are with filter sizes of $3 \times 3$, stride 1, and extract 32-channel feature maps, except for the last convolution layer which outputs a 4-channel final fusion result. The generator can be divided into 3 parts: the embedding network (EmbNet), the fusion network (FunsionNet), and the restoration network (RstNet). EmbNet is a two-stream architecture for feature extraction. It takes a pair of PAN and LR MS images as input and outputs their feature embeddings. The embedding network and restoration network share a similar architecture, both of them consist of one  Block network (RCA Block \cite{RCAGAN} or Residual Block \cite{ResNet}) and one additional convolution layer. The Residual Block is widely used for training deeper neural networks, which consists of two $3 \times 3$ convolution layers with a skip connection. The RCA Block introduces the channel attention mechanism to the residual block and becomes more effective. The detailed influence of them will be discussed in Section~\ref{Block Influence}. The fusion network consists of three Block networks and one additional convolution layer to strengthen its fusion ability. We use the ReLU activation in the generator.

\subsubsection{Discriminator}
Our discriminator follows the architecture of pix2pix \cite{pix2pix} which consists of 5 convolution layers. Strides of the first 3 layers are set to 2 and the last 2 layers are set to 1. All of them are equipped with $4 \times 4$ filters, and the numbers of extracted feature maps in each layer are set to 64, 128, 256, 512, and 1, respectively. The LeakyReLU activation is applied in the discriminator. We use instance normalization \cite{InstanceNorm} in both the generator and discriminator for stable training and better performance.

\subsection{Loss Function}
For training our generator on the full-scale images without ground truths, we introduce a novel hybrid loss function. Our full objective function of the generator can be formulated as follow:
\begin{equation}
	\mathcal{L}_G = \lambda_1 \mathcal{L}_{cyc} + \lambda_2 \mathcal{L}_{adv} + \lambda_3 \mathcal{L}_{rec} + \lambda_4 \mathcal{L}_{nrf}
\end{equation}
The loss function consists of four terms, $i.e.$, cycle loss, adversarial loss,  reconstruction loss, and non-reference loss. $\lambda_1, \lambda_2, \lambda_3, \lambda_4$ are weights to balance the contribution of these loss terms. Next, we introduce them in order.

\begin{figure}[t]
  \centering
  \includegraphics[width=\linewidth]{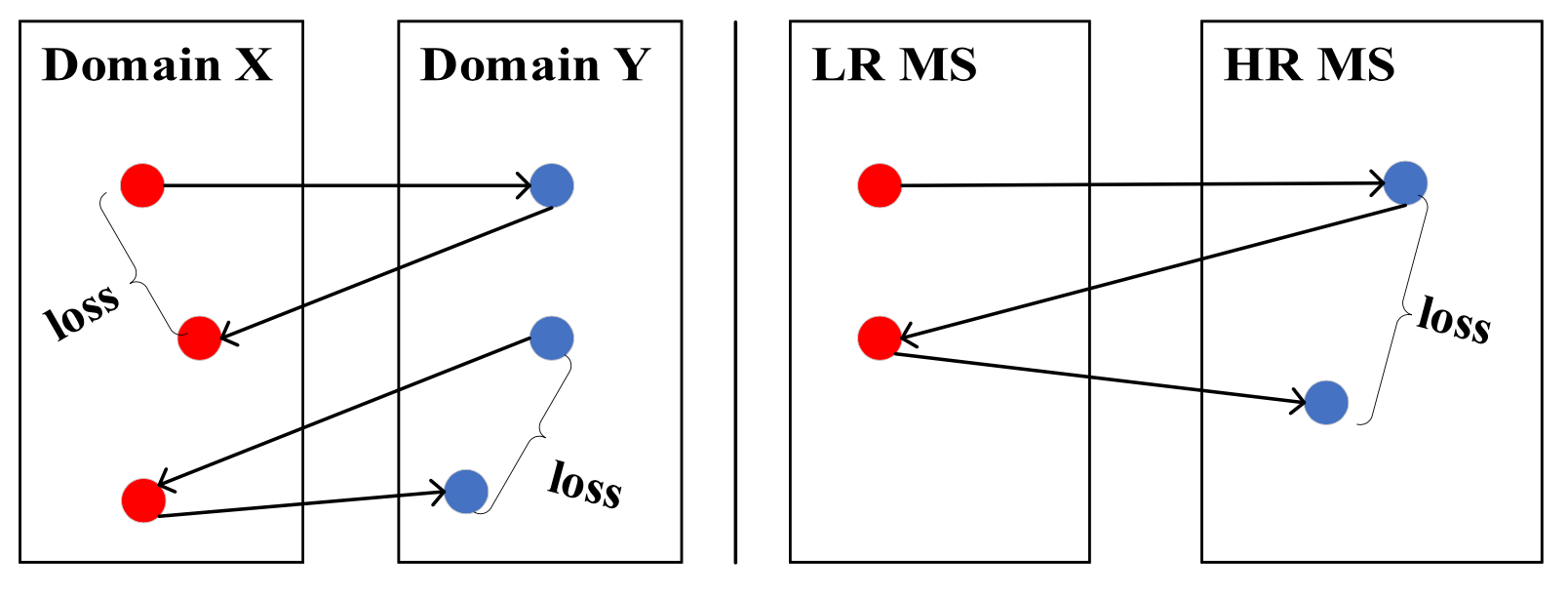}
  \caption{The image on the left shows the traditional cycle-consistency loss in CycleGAN \cite{CycleGAN2017} which is used for translation from domain X to domain Y. The image on the right describes our modified cycle-consistency loss for translating the LR MS image to the HR MS image with the help of the PAN image.}
  \label{fig::loss_cyc}
\end{figure} 

\subsubsection{Cycle Loss}

The concept of cycle-consistency is first introduced by CycleGAN \cite{CycleGAN2017},  which aims to translate an image from a source domain to a target domain in the absence of paired examples. To prevent generators from mode collapse and improve the performance with unpaired images, they meanwhile learn an inverse mapping from the target domain to the source domain and use the cycle loss to force the generators to be each other’s inverse. The left image in Fig.~\ref{fig::loss_cyc} shows the cycle consistency in CycleGAN. $G_y$ maps the image from domain $X$ to domain $Y$ while $G_x$ makes the inverse mapping. The cycle loss requires that for each image $x$ from domain $X$, the image translation cycle should be able to bring $x$ back to the original image, i.e., $x$ should be the same as $G_x(G_y(x))$. And it can be defined similarly in the direction from domain $Y$ to domain $X$. 

Here, in pan-sharpening, we translate the MS image from the LR domain to the HR domain with the help of HR PAN images. Our generator serves a role of mapping the MS image from the LR domain to the HR domain. However, different from CycleGAN where both the images from domain X and domain Y are provided, there are no ideal HR MS images in the real world. To address the MS features being embedded in the HR domain, we translate the LR MS image to the HR MS image twice and check the consistency between them. As the right image in Fig.~\ref{fig::loss_cyc} shows, images are translated to the HR domain, degraded back to the LR domain, and translated to the HR domain again. Here, we choose bicubic interpolation to degrade the images spatially to ease the training process. Compared to CycleGAN, we take one more step forward in the image translation cycle. In this way, if the fusion result loses the information from the MS image, it will cause more differences in the second translation. Our new designed cycle loss $\mathcal{L}_{cyc}$ can be formulated as:

\begin{equation}
\begin{aligned}
	\mathcal{L}_{cyc} = \| G(X, Y) - G(G(X, Y) \downarrow , Y)\|_1
\end{aligned}
\end{equation}	
where $X, Y$ denote the input LR MS and PAN images respectively, $G$ is the two-stream generator, $\downarrow$ means the down-sample operator which degrades the HR image back to the LR image, and $\| \cdot \|_1$ is the $\ell 1$ norm. 

\begin{figure}[t]
  \centering
  \includegraphics[width=\linewidth]{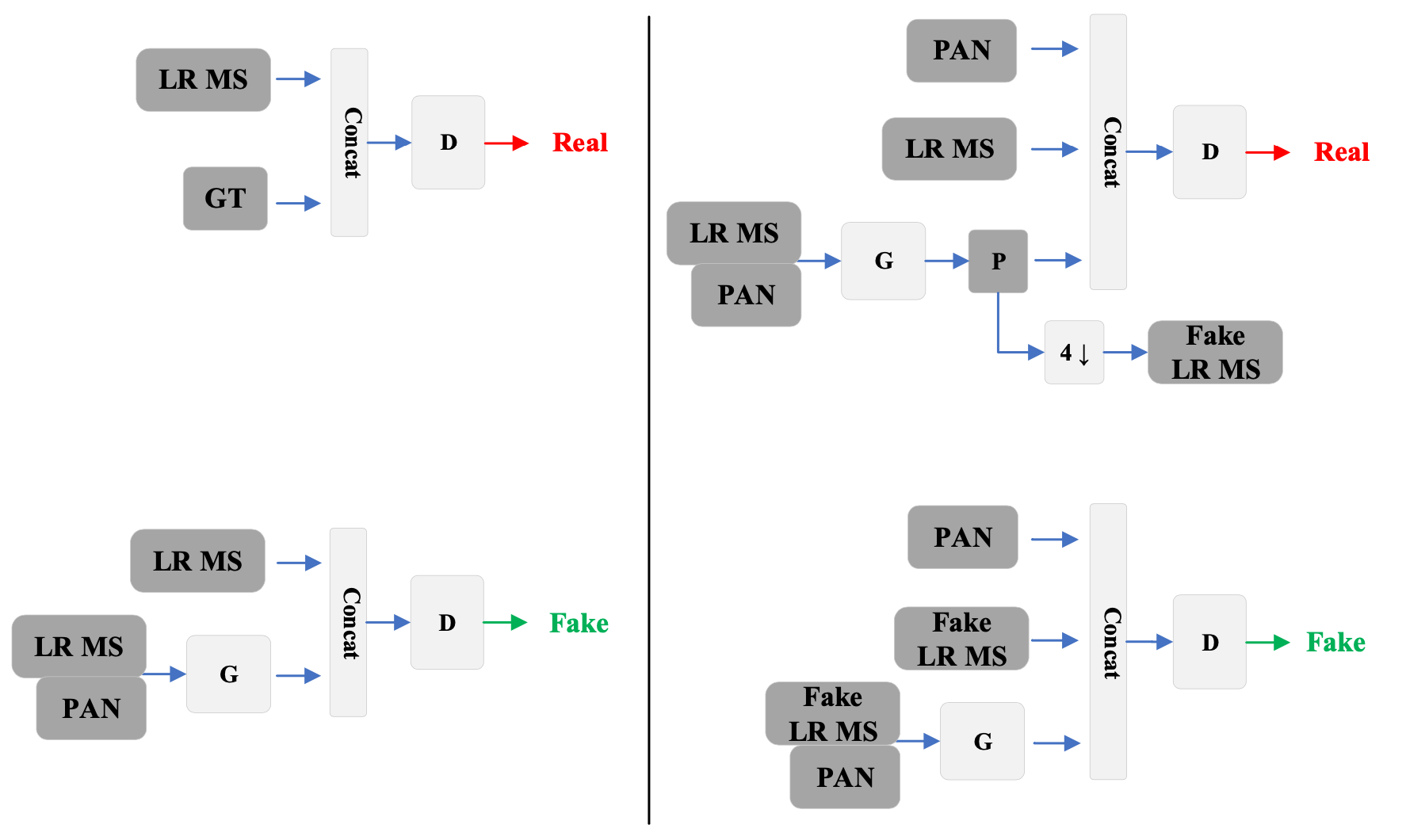}
  \caption{The image on the left shows the adversarial loss in PSGAN \cite{PSGAN} where the discriminator is trained to discriminate the reference MS from the generated pan-sharpened image in the condition of the input LR MS. The image on the right describes our designed adversarial loss where the discriminator needs to distinguish the fake LR MS and its pan-sharpened version in the condition of the PAN.}
  \label{fig::loss_adv}
\end{figure} 
\begin{table*}[!htbp]
\centering
  \caption{Details of the datasets used in the experiments. We build down-scaled and full-scale datasets on GaoFen-2 and WorldView-3 satellites, respectively. Training patches are randomly sampled from the raw remote sensing images while testing patches are cropped in order.}
  \label{tab::dataset}
  \begin{tabular}{ccccrr}
    \toprule
    \textbf{Sensor} & \textbf{bit depth} & \textbf{$\sharp$Images} & \textbf{Scale} & \multicolumn{1}{c}{\textbf{Training}} & \multicolumn{1}{c}{\textbf{Testing}} \\
    \midrule
    \multirow{8}{*}{GaoFen-2} & \multirow{8}{*}{10} & \multirow{8}{*}{7}
    & \multirow{4}{*}{Down} & \textbf{$\sharp$Patches:} 24000 ~~~~~~~~~~~ & \textbf{$\sharp$Patches:} 286 ~~~~~~~~~~~~~~ \\
    & & & & \textbf{PAN:} 256 $\times$ 256 $\times$ 1 & \textbf{PAN:} 400 $\times$ 400 $\times$ 1 \\
    & & & & \textbf{LR MS:} 64 $\times$ 64 $\times$ 4 & \textbf{LR MS:} 100 $\times$ 100 $\times$ 4 \\
    & & & & \textbf{Output:} 256 $\times$ 256 $\times$ 4 & \textbf{Output:} 400 $\times$ 400 $\times$ 4 \\
    \cline{4-6}
    & & & \multirow{4}{*}{Full}  & \textbf{$\sharp$Patches:} 24000 ~~~~~~~~~~~ & \textbf{$\sharp$Patches:} 286 ~~~~~~~~~~~~~~ \\
    & & & & \textbf{PAN:} 256 $\times$ 256 $\times$ 1 & \textbf{PAN:} 400 $\times$ 400 $\times$ 1  \\
    & & & & \textbf{LR MS:} 64 $\times$ 64 $\times$ 4 & \textbf{LR MS:} 100 $\times$ 100 $\times$ 4 \\
    & & & & \textbf{Output:} 256 $\times$ 256 $\times$ 4 &  \textbf{Output:} 400 $\times$ 400 $\times$ 4 \\
    \midrule
    \multirow{8}{*}{WorldView-3} & \multirow{8}{*}{11} & \multirow{8}{*}{3}
    & \multirow{4}{*}{Down} & \textbf{$\sharp$Patches:} 20000 ~~~~~~~~~~~ & \textbf{$\sharp$Patches:} 308 ~~~~~~~~~~~~~~ \\
    & & & & \textbf{PAN:} 256 $\times$ 256 $\times$ 1 & \textbf{PAN:} 400 $\times$ 400 $\times$ 1 \\
    & & & & \textbf{LR MS:} 64 $\times$ 64 $\times$ 4 & \textbf{LR MS:} 100 $\times$ 100 $\times$ 4 \\
    & & & & \textbf{Output:} 256 $\times$ 256 $\times$ 4 & \textbf{Output:} 400 $\times$ 400 $\times$ 4 \\
    \cline{4-6}
    & & & \multirow{4}{*}{Full}  & \textbf{$\sharp$Patches:} 20000 ~~~~~~~~~~~ & \textbf{$\sharp$Patches:} 308 ~~~~~~~~~~~~~~ \\
    & & & & \textbf{PAN:} 256 $\times$ 256 $\times$ 1 & \textbf{PAN:} 400 $\times$ 400 $\times$ 1  \\
    & & & & \textbf{LR MS:} 64 $\times$ 64 $\times$ 4 & \textbf{LR MS:} 100 $\times$ 100 $\times$ 4 \\
    & & & & \textbf{Output:} 256 $\times$ 256 $\times$ 4 &  \textbf{Output:} 400 $\times$ 400 $\times$ 4 \\
  	\bottomrule
\end{tabular}
\end{table*}

\subsubsection{Adversarial Loss}

PSGAN \cite{PSGAN} is one of the first attempts at producing pan-sharpened images with GANs. In addition to the generator, a conditional discriminator network is trained simultaneously to discriminate the reference MS images from the generated pan-sharpened images. The left image in Fig.~\ref{fig::loss_adv} describes the adversarial loss in PSGAN. However, training on the down-scaled images with the synthetic ground truth may cause a gap when dealing with the real world image with the original full-scale resolution.

To overcome the absence of the reference image in the unsupervised setting, similar to our cycle loss, our adversarial loss is also based on the image translation cycle. In the translation cycle, the HR MS image is produced twice, and the first one serves as the ``ground truth''. The pan-sharpened image is first generated by the $G$ using the real LR MS and PAN input and then is degraded to serve as the fake LR MS input using bicubic interpolation for the second generation. $G$ take the fake LR MS image as input and output the fake pan-sharpened result. Therefore, the discriminator is aimed to distinguish the real and fake LR MS and pan-sharpened images in the condition of the input PAN image, while $G$ tries to minimize the differences between the real and fake groups which is also consistent with the aim of $G$ in our cycle loss. The image on the right of Fig.~\ref{fig::loss_adv} depicts our adversarial loss in detail. For stable training and a better performance, we build the adversarial loss based on the least-square loss \cite{LSGAN}. The adversarial loss term in $G$ can be formulated as:

\begin{equation}
	\mathcal{L}_{adv} = ( D(Y, G(X, Y) \downarrow , G(G(X, Y) \downarrow, Y)) - 1) ^ 2 
\end{equation}
where $X, Y$ denote the input LR MS and PAN images respectively, $G$ is our two-stream generator, $D$ stands for the discriminator, $\downarrow$ means the down-sample operator which degrades the pan-sharpened result back to the LR resolution.

Similarly, we can formulate the adversarial loss for $D$:

\begin{equation}
\begin{aligned}
	\mathcal{L}_{D} = ( D(Y, X, G(X, Y)) - a) ^ 2~~~~~~~~~~~~~~~~~~~~~~~~~\\
	+ ( D(Y, G(X, Y) \downarrow , G(G(X, Y) \downarrow, Y)) - b) ^ 2 
\end{aligned}
\end{equation}
where $a$ and $b$ are the soft labels near $1$ and $0$ respectively to avoid the overfitting in discriminator $D$.

\subsubsection{Reconstruction Loss} \label{section::Reconstruction Loss}

Both our cycle loss and adversarial loss mainly focus on the differences in the target domain, i.e., the desired HR MS domain. In addition, for better spatial and spectral preservation, we also add constraints in the LR MS and PAN domains. Specifically, the input LR MS and PAN images are reconstructed from the pan-sharpened output, and then the model evaluates the consistency with the original inputs. Therefore, our reconstruction loss consists of two parts, the spatial part and spectral part:

\begin{equation}
	\mathcal{L}_{rec} = \mathcal{L}_{spatial} + \mathcal{L}_{spectral}
\end{equation}
For spatial preservation, we focus on the high-pass information in the PAN images because the high-pass part generally contains more spatial details and less spectral information. 

\begin{equation} \label{eq::L_spatial}
	\mathcal{L}_{spatial} = \| gethp(Y) - MP(gethp(G(X, Y)))\|_1
\end{equation} 
$X, Y$ denote the input LR MS and PAN images, respectively, $G$ is our two-stream generator, $gethp(\cdot)$ extracts the high-pass information from the image, and $MP(\cdot)$ is the maximum pooling operation for converting pan-sharpened image into a single channel while keeping the details.

As for the spectral preservation, we focus on the low-pass information of the MS images because the low-pass part generally contains more spectral information and less spatial details.
 
\begin{equation}
	\mathcal{L}_{spectral} = \| getlp(X) - getlp(G(X, Y) \downarrow )\|_1
\end{equation}
where $getlp(\cdot)$ extracts the low-pass information from the image, and $\downarrow$ is the down-sampling operation for converting pan-sharpened image into the same resolution as the input LR MS image. We use the same method for the high-pass and low-pass extraction as \cite{PanNet} for its great progress in the image fusion field.

\subsubsection{Non-Reference Loss}
Following \cite{pgman}, we introduce Q-loss to boost performance of our model. It is defined as follows:

\begin{equation}
	\mathcal{L}_{nrf} = 1 - QNR
\end{equation}

where QNR \cite{QNR} is a non-reference image quality assessment metric. 

\subsection{Training Details}
We implement our proposed method in PyTorch \cite{PyTorch} and train it on a single NVIDIA Titan 2080Ti GPU. The batch size and learning rate are set as 8 and $ 1 \times 10^{-4}$, respectively. The trade-off parameters in our loss function are set as $\lambda_1 = 1 \times 10^{-3}, \lambda_2 = 1 \times 10^{-3}, \lambda_3 = 5 \times 10^{-4}$, and $\lambda_4 = 1$. The soft label $a$ is a random number ranging from $0.7$ to $1.2$ and the soft label $b$ is  a random number ranging from $0$ to $0.3$. Adam optimizer \cite{Adam} is used to train the model for 20 epochs with fixed hyper-parameters $\beta_1 =0.5$ and $\beta_2=0.999$. In addition, all images are processed at their raw bit depth in both training and testing stages to be the same as the condition in the real world.  

\begin{table*}[t]
\centering
  \caption{Ablation study of the loss function on GaoFen-2 dataset. ``\checkmark" means including the referred loss term. The last row indicates the ideal value of each metric.}
  \label{tab::loss ablation}
  \begin{tabular}{c|c|c|c||c|c|c|c|c|c}
    \toprule
    \multicolumn{4}{c||}{Loss Term} & \multicolumn{3}{c|}{Non-reference Metrics} & \multicolumn{3}{c}{Reference Metrics} \\
    $\mathcal{L}_{cyc}$ & $\mathcal{L}_{adv}$ & 	$\mathcal{L}_{rec}$ &  $\mathcal{L}_{nrf}$	& ~~D$_\lambda$~~ & ~~D$_s$~~ & ~~QNR~~ & ~~SAM~~ & ERGAS & ~SSIM~ \\
    \midrule
	\checkmark & \checkmark & \checkmark & \checkmark & 0.005 & 	0.013 & 0.982 & 1.680 &	2.155 &	0.935 \\ 
	\midrule
			   & \checkmark & \checkmark & \checkmark	& 0.007	& 0.013	& 0.980	& 1.952	 & 2.474	& 0.923 \\
	\checkmark & 			& \checkmark & \checkmark 	& 0.006	& 0.013	& 0.981	& 1.676	& 2.156	& 0.935 \\
	\checkmark & \checkmark & 			 & \checkmark 	& 0.006	& 0.012	& 0.982	& 1.622	& 2.225	& 0.928 \\
	\checkmark & \checkmark & \checkmark & 				& 0.040	& 0.117	& 0.849	& 3.444	& 3.636	& 0.827 \\
	\midrule
	\checkmark & 			& 			 & 				& 0.001	& 0.129	& 0.870 & 2.080 & 3.666	& 0.815 \\
			   & \checkmark & 			 & 				& 0.356	& 0.376	& 0.405	& 3.615	& 10.123 & 0.776 \\
			   & 			& \checkmark & 				& 0.048	& 0.115	& 0.843	& 3.421	& 3.589	& 0.831 \\
			   & 			& 			 & \checkmark 	& 0.005	& 0.012	& 0.983	& 2.341	& 2.800	& 0.916 \\ 
	\midrule
	\multicolumn{4}{c||}{Ideal Value}  & 0	& 0 	& 1	& 0	& 0 	& 1\\
	\bottomrule
\end{tabular}
\end{table*}

\section{Experiments and Results} \label{section::Experiments and Results}

\subsection{Datasets}
We conduct extensive experiments on two image sets collected from GaoFen-2 and WorldView-3 satellites, respectively. For comparison with other methods, we report results under both Wald's protocol~\cite{wald} and full-scale setting. Following Wald's protocol, both the PAN and MS images are down-sampled to a lower scale so that the original MS images serve as the ground truth images. This setting is widely used existing fusion literatures. Same as most of the related works \cite{PanNet, Pan-GAN}, we blur the original image pairs using a Gaussian filter and then down-sample them with a scaling factor of 4. Under full-scale setting, there are no reference images, models are trained in an unsupervised manner. Because the size of remote sensing images is so large that it is difficult to feed them into neural networks, we crop these images into small patches to form training and testing sets. At first, we crop testing images into $286$ patches to build the testing set, while the training set consisting of $24,000$ patches are randomly cropped from other images. Then we conduct the same process on WorldView-3 dataset which contains $308$ testing patches and $20,000$ training patches. In training process we feed the model with PAN images in size of $256 \times 256 \times 1$ and LR MS images in size of $64 \times 64 \times 4$, while the sizes of PAN and LR MS images are $400 \times 400 \times 1$ and $100 \times 100 \times 4$ respectively in testing. The patches keep the same bit depth as the raw remote sensing images without any other pre-processing, so models need to learn the original distributions of PAN and MS images.

To perform the evaluation on full-scale images, we employ 3 widely used non-reference metrics: $\bm{D_\lambda}$ \cite{QNR}, $\bm{D_s}$ \cite{QNR} and QNR \cite{QNR}. Evaluations under Wald's protocol \cite{wald} adopt the following reference metrics: spectral angle mapper (SAM) \cite{SAM}, relative dimensionless global error in synthesis (ERGAS) \cite{ERGAS} and SSIM \cite{SSIM}.

\begin{table*}[t]
\centering
  \caption{Parameter sensitive test for our proposed model on GaoFen-2 Dataset. The last row indicates the ideal value of each metric.}
  \label{tab::params}
  \begin{tabular}{c|c|c|c||c|c|c|c|c|c}
    \toprule
    \multicolumn{4}{c||}{Weight} & \multicolumn{3}{c|}{Non-reference Metrics} & \multicolumn{3}{c}{Reference Metrics} \\
    ~~~$\lambda_1$~~~ & ~~~$\lambda_2$~~~ & ~~~$\lambda_3$~~~ &  ~~~$\lambda_4$~~~ 	& ~~D$_\lambda$~~ & ~~D$_s$~~ & ~~QNR~~ & ~~SAM~~ & ERGAS & ~SSIM~ \\
    \midrule
    0.0001	& 0.001		& 0.0005 	& 1		& 0.007	& 0.012	& 0.981	& 1.790	& 2.240	& 0.931 \\
	\textbf{0.001}	& 0.001		& 0.0005	 	& 1		& 0.005 & 	0.013 & 0.982 & 1.680 &	2.155 &	0.935\\
	0.01	& 0.001		& 0.0005 	& 1		& 0.007	& 0.015	& 0.978	& 1.731	& 2.569	& 0.910 \\
	\midrule
	0.001	& 0.0001 	& 0.0005 	& 1 		& 0.005	& 0.013	& 0.982	& 1.668	& 2.151	& 0.936 \\
	0.001	& \textbf{0.001}		& 0.0005	 	& 1		& 0.005 & 	0.013 & 0.982 & 1.680 &	2.155 &	0.935 \\
	0.001	& 0.01		& 0.0005		& 1		& 0.011	& 0.015	& 0.975	& 1.972	& 2.641	& 0.911 \\
	\midrule
	0.001	& 0.001		& 0.00005	& 1		& 0.007 & 0.013	& 0.981	& 1.617	& 2.166	& 0.932 \\
	0.001	& 0.001		& \textbf{0.0005}		& 1		& 0.005 & 	0.013 & 0.982 & 1.680 &	2.155 &	0.935	\\
	0.001	& 0.001		& 0.005		& 1		& 0.007	& 0.02	& 0.974	& 2.034	& 2.551	& 0.904 \\
	\midrule
	0.001	& 0.001		& 0.0005		& 0.1	& 0.007	& 0.022	& 0.972	& 2.042	& 2.591	& 0.909 \\
	0.001	& 0.001		& 0.0005		& \textbf{1}		& 0.005 & 	0.013 & 0.982 & 1.680 &	2.155 &	0.935 \\
	0.001	& 0.001		& 0.0005		& 10		& 0.006	& 0.014	& 0.980	& 1.848	& 2.354	& 0.926 \\
    \midrule
	\multicolumn{4}{c||}{Ideal Value}  & 0	& 0 	& 1	& 0	& 0 	& 1\\
	\bottomrule
\end{tabular}
\end{table*}

\begin{table*}[t]
\centering
  \caption{Extended ablation study on GaoFen-2 dataset. ''UCGAN`` is our proposed model, ''UCGAN-Res`` is the variant using with residual blocks, and ''UCGAN-AP`` uses average pooling (AP) when converting pan-sharpened results to PAN images. The last row indicates the ideal value of each metric.}
  \vspace{2pt}
  \label{tab::extended ablation}
  \begin{tabular}{l||c|c|c|c|c|c}
    \toprule
    \multirow{2}{*}{Model} & \multicolumn{3}{c|}{Non-reference Metrics} & \multicolumn{3}{c}{Reference Metrics} \\
    & ~~D$_\lambda$~~ & ~~D$_s$~~ & ~~QNR~~ & ~~SAM~~ & ERGAS & ~SSIM~ \\
    \midrule 
	UCGAN	& 0.005 & 	0.013 & 0.982 & 1.680 &	2.155 &	0.935 \\ 
	UCGAN-AP & 0.006 & 0.013	& 0.981	& 1.697	& 2.160	& 0.936 \\
	UCGAN-Res & 0.009 &	 0.015 & 	0.977 &	1.714 &	2.184 &	0.933 \\
	\midrule
	Ideal Value & 0	& 0 	& 1	& 0	& 0 & 1 \\
	\bottomrule
\end{tabular}
\end{table*}

\subsection{Ablation Study}

\subsubsection{Loss Terms}
In this paper, we introduce a novel loss function which consists of four loss terms. To verify the impact and performance of each component in our loss function, we design 8 variants of our model: 4 for abandoning one of the loss terms and another 4 for only using one of the loss terms. All variants are trained and tested in the same scheme except for the loss function. Table~\ref{tab::loss ablation} shows the result of the ablation study on GaoFen-2 dataset. When removing the cycle loss from our model, the reference metrics including SAM, ERGAS, and SSIM decrease dramatically. It is demonstrated that the cycle-consistency is quite important across scales and helps the model perform well on the down-scaled images. When training without the adversarial loss, indexes decrease slightly. Generally, the main contribution of the adversarial learning is to improve the visual perception quality while little improvement for quantitative results. When the reconstruction loss is excluded, there is a clear drop in most metrics. When removing the non-reference loss, the performance drops significantly. On the other hand, if only using one of these loss terms, nearly all metrics fall a lot. It is concluded that using the full loss function is the optimal way to make the model to achieve its best performance. We also give a detailed discussion about the hyper-parameter setting of the loss function in Table~\ref{tab::params}. For each weight variable in the loss function, we have tried 3 different values. It is observed that our model is not so sensitive to the hyper-parameters even if we change them to different values. We choose the best performance among these combinations as our final parameter setting.

\subsubsection{RCA Block vs Residual Block} \label{Block Influence}

As Fig.~\ref{fig::architecture} describes, we have two options of the Block used in the generator, i.e., the residual block \cite{ResNet} and the RCA Block \cite{RCAGAN}. The Residual Block is first proposed in ResNet \cite{ResNet}, which consists of two $3 \times 3$ convolution layers with a skip connection. Residual blocks make it possible to train deeper neural networks without suffering from gradient vanishing problem and are widely used in many fields. On the other hand, to exploit the interdependencies among different feature channels, Cai \etal \cite{RCAGAN} combine the channel attention mechanism with the residual block and propose the Residual-Channel-Attention (RCA) block. The main idea of RCA block is that they use a global average pooling layer to squeeze each input channel into a channel descriptor and then feed them into two fully-connected (FC) layers to produce channel-wise scaling factors for the input channel. After modeling interdependencies between different channels, RCA block retains the same dimension as the input channel which is very flexible and can be easily applied to our model. To examine the superiority of these two architectures, we design two kinds of generators based on the two blocks and conduct the ablation study on GaoFen-2 dataset. From Table~\ref{tab::extended ablation}, we can observe that replacing the RCA block with the residual block will cause corruption on non-reference metrics, which may distort our pan-sharpened results on full-scale images. Therefore, we finally employ the RCA Block to further improve the performance of our model.

\subsubsection{Maximum Pooling vs Average Pooling}

In section \ref{section::Reconstruction Loss}, we add the spatial constraint on our model by reconstructing the PAN image from the pan-sharpened image for better preservation of spatial details. As Eq~(\ref{eq::L_spatial}) shows, we need to make comparisons between the PAN image, a single-channel image, and the pan-sharpened result, a 4-channel image. A similar situation is also faced by \cite{Pan-GAN}. They use an average pooling operation to obtain the intensity image of the pan-sharpened result. However, calculating the mean value of an image among the channel dimension may blur the image which is harmful for spatial preservation. To alleviate this issue, we consider the maximum pooling operation instead. The maximum pooling operation can preserve sharper spatial details and reduce the blurring effect. To validate our assumption, we compare our model with a variant one that use the average pooling, termed as ''UCGAN-AP``. We test the modified model on GaoFen-2 dataset, and the results are displayed in Table~\ref{tab::extended ablation}.

\begin{table*}[t]
\centering
  \caption{Quantitative results of GaoFen-2 dataset. The best result in each group is in \textbf{bold} font. The last row indicates the ideal value of each metric.}
  \label{tab::GF-2 result}
  \begin{tabular}{r|r||c|c|c|c|c|c}
    \toprule
    \multirow{2}{*}{Type} & \multirow{2}{*}{Model}	& \multicolumn{3}{c|}{Non-reference Metrics} & \multicolumn{3}{c}{Reference Metrics} \\
     & & ~~D$_\lambda$~~ & ~~D$_s$~~ & ~~QNR~~ & ~~SAM~~ & ERGAS & ~SSIM~ \\
    \midrule
    \multirow{7}{*}{Traditional} & GS \cite{GS}	& 0.030$\pm$0.025 		& 0.048$\pm$0.032	& 0.925$\pm$0.051	& 1.967$\pm$0.205	& 3.719$\pm$0.375 	& 0.838$\pm$0.018 \\
    & IHS \cite{IHS} & 0.028$\pm$0.022		& 0.045$\pm$0.032	& 0.929$\pm$0.050	& 2.130$\pm$0.215	& 3.859$\pm$0.402	& 0.838$\pm$0.018 \\
    & Brovey \cite{Brovey} & 0.025$\pm$0.017		& 0.051$\pm$0.030	& 0.927$\pm$0.042	& \textbf{1.737$\pm$0.205} 	& 3.433$\pm$0.320	& 0.829$\pm$0.017 \\
    & HPF \cite{HPFandHPFC}	& 0.020$\pm$0.012		& \textbf{0.018$\pm$0.008}	& 0.962$\pm$0.018	& 2.083$\pm$0.231	& 4.105$\pm$0.448	& 0.810$\pm$0.020 \\
    & LMM \cite{LMVMandLMM}	& 0.022$\pm$0.014		& 0.025$\pm$0.014	& 0.954$\pm$0.026	& 1.814$\pm$0.223	& 6.337$\pm$0.683	& 0.640$\pm$0.027 \\
    & LMVM \cite{LMVMandLMM} 	& \textbf{0.004$\pm$0.002}		& 0.099$\pm$0.034	& 0.898$\pm$0.035	& 1.863$\pm$0.215	& \textbf{2.715$\pm$0.330}	& \textbf{0.886$\pm$0.015} \\
    & SFIM \cite{SFIM} & 0.018$\pm$0.011		& 0.019$\pm$0.007	& \textbf{0.963$\pm$0.017}	& 1.951$\pm$0.188	& 4.261$\pm$0.486	& 0.809$\pm$0.021 \\
    \midrule
    \multirow{5}{*}{Supervised} & PNN \cite{PNN} & 0.015$\pm$0.011		& \textbf{0.031$\pm$0.014}	& \textbf{0.954$\pm$0.015}	& 1.272$\pm$0.116	& 1.498$\pm$0.200	& 0.963$\pm$0.006 \\
    & DRPNN \cite{DRPNN} & 0.017$\pm$0.009		& 0.070$\pm$0.029	& 0.915$\pm$0.031	& 1.216$\pm$0.120	& 1.284$\pm$0.151	& 0.971$\pm$0.004 \\
    & MSDCNN \cite{MSDCNN} & 0.013$\pm$0.008 		& 0.041$\pm$0.023	& 0.947$\pm$0.023	& 1.442$\pm$0.154	& 1.490$\pm$0.207	& 0.963$\pm$0.006 \\
    & PanNet \cite{PanNet} & 0.011$\pm$0.007		& 0.123$\pm$0.043	& 0.867$\pm$0.046	& 1.088$\pm$0.114	& 1.292$\pm$0.166	& 0.971$\pm$0.005 \\
    & PSGAN \cite{PSGAN} 	& \textbf{0.008$\pm$0.004}		& 0.101$\pm$0.036	& 0.891$\pm$0.037	& \textbf{1.057$\pm$0.104}	& \textbf{1.209$\pm$0.143}	& \textbf{0.975$\pm$0.004} \\
    \midrule
    \multirow{2}{*}{Unsupervised} & Pan-GAN \cite{Pan-GAN} & 0.035$\pm$0.019		& 0.027$\pm$0.018	& 0.939$\pm$0.034	& 2.642$\pm$0.270	& 5.245$\pm$0.561	& 0.732$\pm$0.024 \\
    & Ours 	& \textbf{0.005$\pm$0.005} &	\textbf{0.013$\pm$0.008} & 	\textbf{0.982$\pm$0.011} & 	\textbf{1.680$\pm$0.188} &	\textbf{2.155$\pm$0.271}	& \textbf{0.935$\pm$0.011} \\
	\midrule
	\multicolumn{2}{c||}{Ideal Value} & 0	& 0 	& 1	& 0	& 0 	& 1 \\
	\toprule
\end{tabular}
\end{table*}

\begin{table*}[t]
\centering
  \caption{Quantitative results of WorldView-3 dataset. The best result in each group is in \textbf{bold} font. The last row indicates the ideal value of each metric.}
  \label{tab::WV-3 result}
  \begin{tabular}{r|r||c|c|c|c|c|c}
    \toprule
    \multirow{2}{*}{Type} & \multirow{2}{*}{Model}	& \multicolumn{3}{c|}{Non-reference Metrics} & \multicolumn{3}{c}{Reference Metrics} \\
     & & ~~D$_\lambda$~~ & ~~D$_s$~~ & ~~QNR~~ & ~~SAM~~ & ERGAS & ~SSIM~ \\
    \midrule
    \multirow{7}{*}{Traditional} & GS \cite{GS}	& 0.028$\pm$0.018		& 0.092$\pm$0.014	& 0.883$\pm$0.021	& 5.480$\pm$2.338	& 7.133$\pm$2.360	& 0.855$\pm$0.119 \\
    & IHS \cite{IHS} & 0.033$\pm$0.019		& 0.085$\pm$0.015	& 0.884$\pm$0.025	& 5.299$\pm$3.078	& 6.935$\pm$2.627	& 0.855$\pm$0.124 \\
    & Brovey \cite{Brovey} & \textbf{0.020$\pm$0.014}		& 0.093$\pm$0.015	& 0.887$\pm$0.021	& 4.351$\pm$3.639	& 6.615$\pm$3.106	& \textbf{0.862$\pm$0.124} \\
    & HPF \cite{HPFandHPFC}	& 0.026$\pm$0.008		& 0.092$\pm$0.018	& 0.884$\pm$0.021	& 4.140$\pm$3.443	& \textbf{6.098$\pm$3.654}	& 0.850$\pm$0.139 \\
    & LMM \cite{LMVMandLMM}	& 0.035$\pm$0.010		& 0.096$\pm$0.020	& 0.873$\pm$0.025	& 4.456$\pm$3.734	& 6.311$\pm$3.793	& 0.852$\pm$0.138 \\
    & LMVM \cite{LMVMandLMM} & 0.021$\pm$0.016		& \textbf{0.070$\pm$0.011}	& \textbf{0.910$\pm$0.022}	& 4.540$\pm$3.807	& 7.451$\pm$4.568	& 0.776$\pm$0.214 \\
    & SFIM \cite{SFIM} & 0.021$\pm$0.007		& 0.088$\pm$0.017	& 0.893$\pm$0.019	& \textbf{3.838$\pm$3.154}	& 6.112$\pm$3.683	& 0.854$\pm$0.136 \\
    \midrule
    \multirow{5}{*}{Supervised} & PNN \cite{PNN} & 0.037$\pm$0.021		& \textbf{0.193$\pm$0.026}	& \textbf{0.778$\pm$0.032}	& 3.678$\pm$2.541	& 5.731$\pm$3.035	& 0.890$\pm$0.102 \\
    & DRPNN \cite{DRPNN} & 0.053$\pm$0.029		& 0.264$\pm$0.029	& 0.698$\pm$0.038	& 3.450$\pm$2.496	& \textbf{5.532$\pm$2.936}	& 0.895$\pm$0.097 \\
    & MSDCNN \cite{MSDCNN} & 0.046$\pm$0.029		& 0.237$\pm$0.027	& 0.728$\pm$0.035	& 3.341$\pm$2.532	& 5.569$\pm$3.142	& 0.894$\pm$0.099 \\
    & PanNet \cite{PanNet} & 0.062$\pm$0.030		& 0.257$\pm$0.028	& 0.697$\pm$0.040	& 3.585$\pm$2.670	& 5.833$\pm$3.040	& 0.886$\pm$0.105 \\
    & PSGAN \cite{PSGAN} & \textbf{0.030$\pm$0.019}		& 0.243$\pm$0.027	& 0.735$\pm$0.032	& \textbf{3.118$\pm$2.412}	& 5.603$\pm$3.198	& \textbf{0.898$\pm$0.096} \\
    \midrule
    \multirow{2}{*}{Unsupervised} & Pan-GAN \cite{Pan-GAN} & 0.047$\pm$0.010		& 0.120$\pm$0.024	& 0.839$\pm$0.029	& \textbf{4.504$\pm$3.606}	& \textbf{6.040$\pm$3.256}	& \textbf{0.867$\pm$0.123} \\
    & Ours 	& \textbf{0.016$\pm$0.011}		& \textbf{0.052$\pm$0.009}	& \textbf{0.933$\pm$0.017}	& 4.517$\pm$3.747	& 6.777$\pm$4.041	& 0.819$\pm$0.169 \\
	\midrule
	\multicolumn{2}{c||}{Ideal Value} & 0	 & 0 	& 1	& 0	& 0 	& 1 \\
	\bottomrule
\end{tabular}
\end{table*}

\subsection{Comparative Experiments}

\subsubsection{Comparison Models}
To demonstrate the strength of our model, we compare it with 13 state-of-art pan-sharpening methods on GaoFen-2 and WorldView-3 datasets. These models can be categorized into three groups: traditional methods, supervised learning methods, and unsupervised learning methods.

\begin{itemize}
\item \textbf{Traditional methods} include GS \cite{GS}, IHS \cite{HCS}, Brovey \cite{Brovey}, HPF \cite{HPFandHPFC}, LMM \cite{LMVMandLMM}, LMVM \cite{LMVMandLMM}, and SFIM \cite{SFIM}. These methods can be tested without training and can be applied to both down-scaled and full-scale image sets, directly. 
\item \textbf{Supervised methods} include PNN \cite{PNN}, DRPNN \cite{DRPNN}, MSDCNN \cite{MSDCNN}, PanNet \cite{PanNet}, and PSGAN \cite{PSGAN}. These methods are state-of-the-art supervised deep learning based pan-sharpening methods. They are trained and tested under Wald's protocol. We generalize them to the full-scale images to obtain HR pan-sharpened images.
\item \textbf{Unsupervised methods} include Pan-GAN \cite{Pan-GAN} and ours. Unsupervised models can be trained and tested under both settings, so we report quantitative results on both down-scaled and full-scale pan-sharpened images. 
\end{itemize}

For fair comparisons, we have tried our best to train the learning-based models and selected the optimal performances of them for comparison. 
\begin{figure*}[!htb]
\centering
	\subfigure[]{\label{fig::gf_pan}
		\includegraphics[width=.24\linewidth]{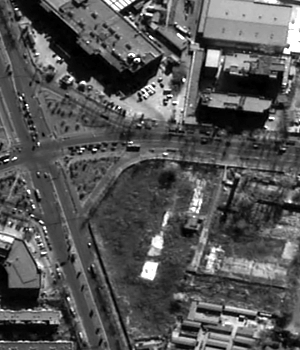}} \hspace{-9pt}
	\subfigure[]{\label{fig::gf_lrms}
		\includegraphics[width=.24\linewidth]{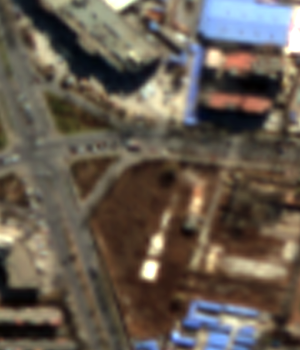}} \hspace{-9pt}
	\subfigure[]{\label{fig::gf_GS}
		\includegraphics[width=.24\linewidth]{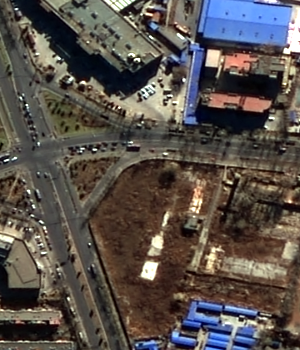}} \hspace{-9pt}
	\subfigure[]{\label{fig::gf_IHS}
		\includegraphics[width=.24\linewidth]{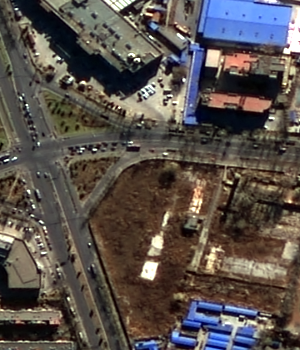}} \\
		\vspace{-8pt}
	\subfigure[]{\label{fig::gf_Brovey}
		\includegraphics[width=.24\linewidth]{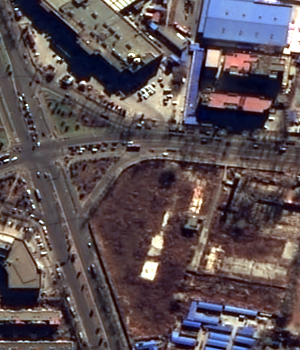}} \hspace{-9pt}
	\subfigure[]{\label{fig::gf_HPF}
		\includegraphics[width=.24\linewidth]{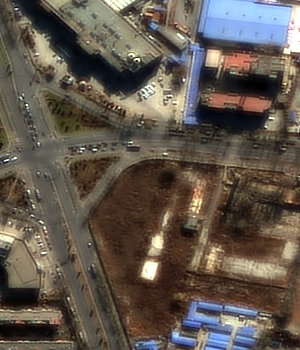}} \hspace{-9pt}
	\subfigure[]{\label{fig::gf_LMM}
		\includegraphics[width=.24\linewidth]{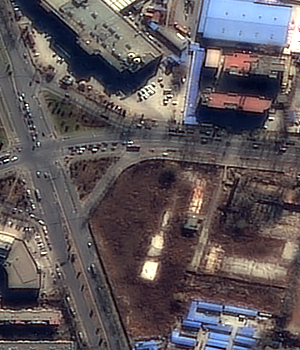}} \hspace{-9pt}
	\subfigure[]{\label{fig::gf_LMVM}
		\includegraphics[width=.24\linewidth]{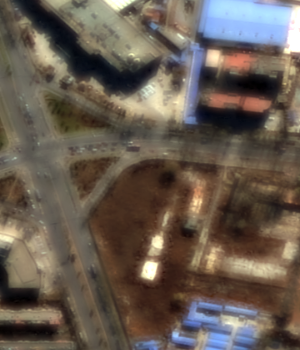}} \\
		\vspace{-8pt}
	\subfigure[]{\label{fig::gf_SFIM}
		\includegraphics[width=.24\linewidth]{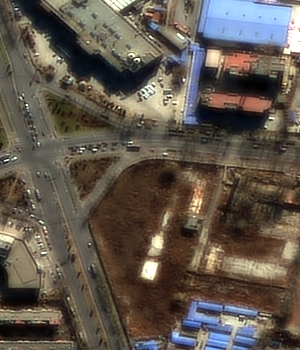}} \hspace{-9pt}
	\subfigure[]{\label{fig::gf_pnn}
		\includegraphics[width=.24\linewidth]{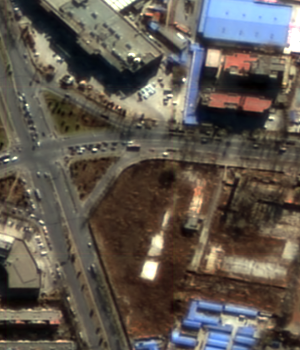}} \hspace{-9pt}
	\subfigure[]{\label{fig::gf_drpnn}
		\includegraphics[width=.24\linewidth]{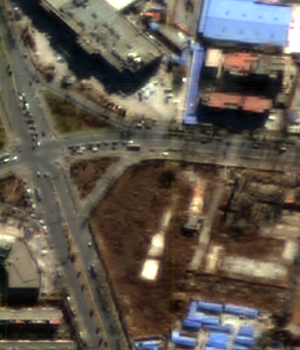}} \hspace{-9pt}
	\subfigure[]{\label{fig::gf_msdcnn}
		\includegraphics[width=.24\linewidth]{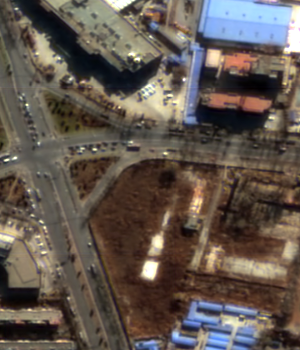}} \\
		\vspace{-8pt}
	\subfigure[]{\label{fig::gf_pannet}
		\includegraphics[width=.24\linewidth]{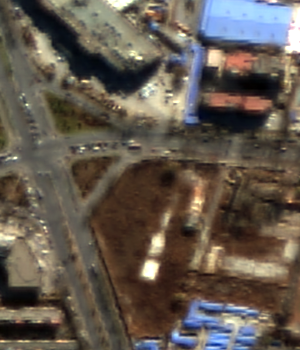}} \hspace{-9pt}
	\subfigure[]{\label{fig::gf_psgan}
		\includegraphics[width=.24\linewidth]{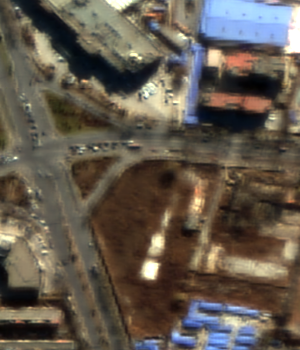}} \hspace{-9pt}
	\subfigure[]{\label{fig::gf_pan_gan-2}
		\includegraphics[width=.24\linewidth]{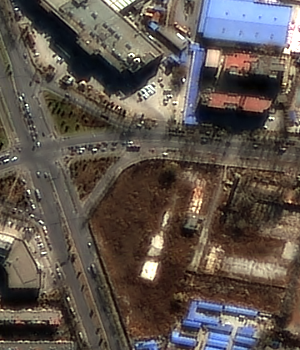}} \hspace{-9pt}
	\subfigure[]{\label{fig::gf_model90-2}
		\includegraphics[width=.24\linewidth]{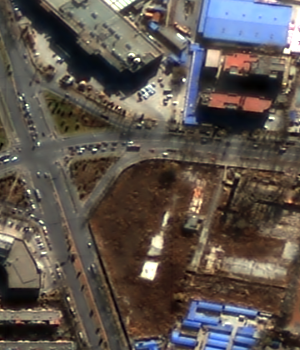}} \\	
		\vspace{-8pt}
	\caption{Pansharpening results on the GaoFen-2 full-scale images. Visualized in RGB. (a) PAN. (b) LR MS. (c) GS \cite{GS}. (d) IHS \cite{IHS}. (e) Brovey \cite{Brovey}. (f) HPF \cite{HPFandHPFC}. (g) LMM \cite{LMVMandLMM}. (h) LMVM \cite{LMVMandLMM}. (i) SFIM \cite{SFIM}. (j) PNN \cite{PNN}. (k) DRPNN \cite{DRPNN}. (l) MSDCNN \cite{MSDCNN}. (m) PanNet \cite{PanNet}. (n) PSGAN \cite{PSGAN}. (o) Pan-GAN \cite{Pan-GAN}. (p) Ours. }
\label{fig::GF results}
\end{figure*}

\begin{figure*}[!htb]
\centering
	\subfigure[]{\label{fig::wv_pan}
		\includegraphics[width=.24\linewidth]{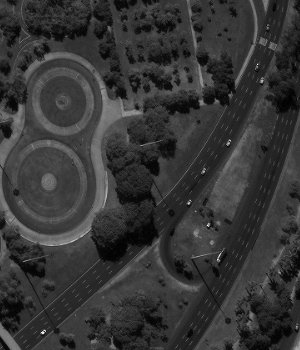}} \hspace{-9pt}
	\subfigure[]{\label{fig::wv_lrms}
		\includegraphics[width=.24\linewidth]{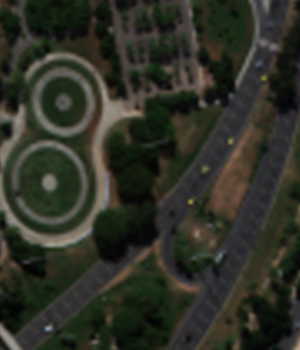}} \hspace{-9pt}
	\subfigure[]{\label{fig::wv_GS}
		\includegraphics[width=.24\linewidth]{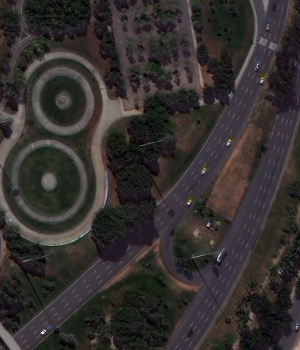}} \hspace{-9pt}
	\subfigure[]{\label{fig::wv_IHS}
		\includegraphics[width=.24\linewidth]{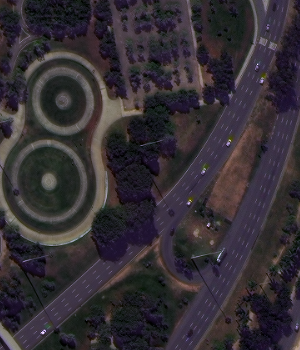}} \\
		\vspace{-8pt}
	\subfigure[]{\label{fig::wv_Brovey}
		\includegraphics[width=.24\linewidth]{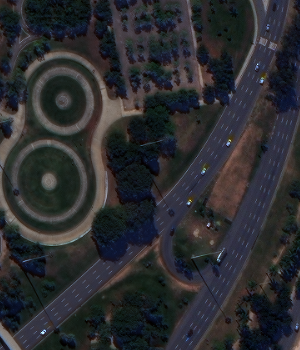}} \hspace{-9pt}
	\subfigure[]{\label{fig::wv_HPF}
		\includegraphics[width=.24\linewidth]{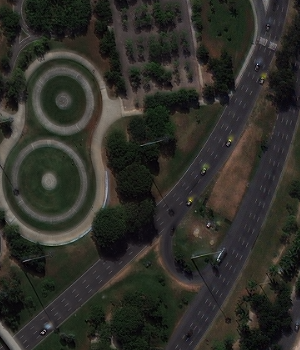}} \hspace{-9pt}
	\subfigure[]{\label{fig::wv_LMM}
		\includegraphics[width=.24\linewidth]{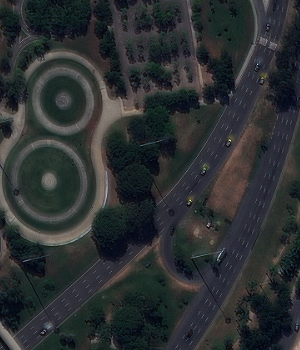}} \hspace{-9pt}
	\subfigure[]{\label{fig::wv_LMVM}
		\includegraphics[width=.24\linewidth]{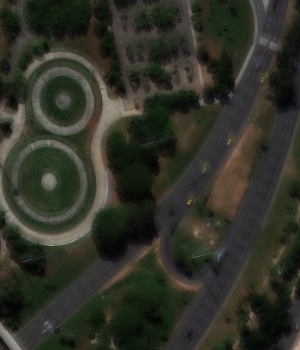}} \\
		\vspace{-8pt}
	\subfigure[]{\label{fig::wv_SFIM}
		\includegraphics[width=.24\linewidth]{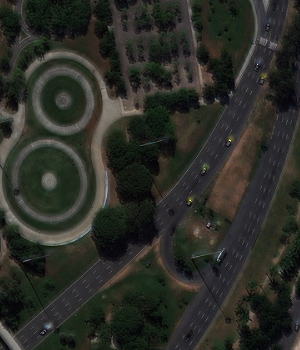}} \hspace{-9pt}
	\subfigure[]{\label{fig::wv_pnn}
		\includegraphics[width=.24\linewidth]{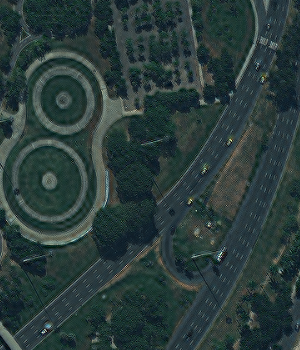}} \hspace{-9pt}
	\subfigure[]{\label{fig::wv_drpnn}
		\includegraphics[width=.24\linewidth]{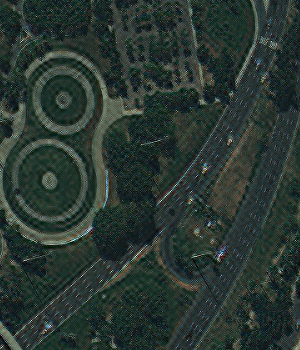}} \hspace{-9pt}
	\subfigure[]{\label{fig::wv_msdcnn}
		\includegraphics[width=.24\linewidth]{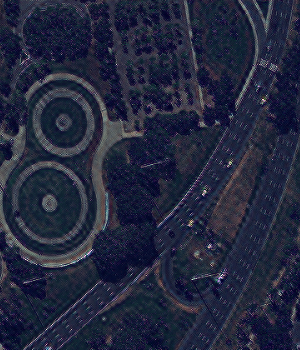}} \\
		\vspace{-8pt}
	\subfigure[]{\label{fig::wv_pannet}
		\includegraphics[width=.24\linewidth]{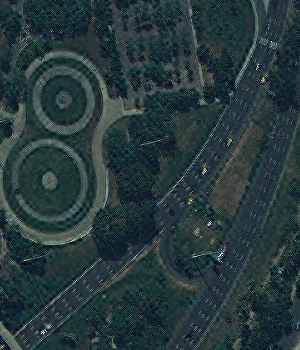}} \hspace{-9pt}
	\subfigure[]{\label{fig::wv_psgan}
		\includegraphics[width=.24\linewidth]{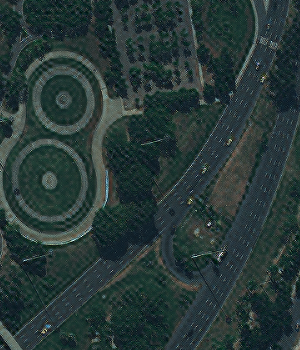}} \hspace{-9pt}
	\subfigure[]{\label{fig::wv_pan_gan-2}
		\includegraphics[width=.24\linewidth]{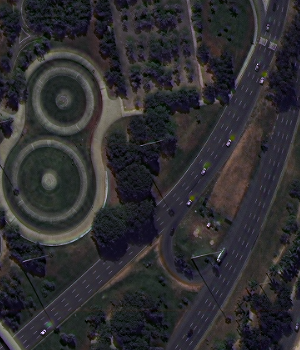}} \hspace{-9pt}
	\subfigure[]{\label{fig::wv_model90-2}
		\includegraphics[width=.24\linewidth]{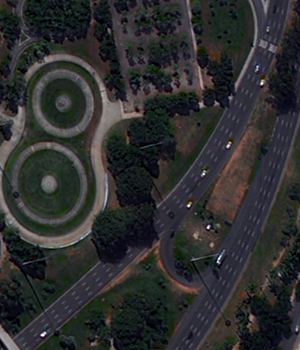}} \\	
		\vspace{-8pt}
	\caption{Pansharpening results on the WorldView-3 full-scale images. Visualized in RGB. Visualized in RGB. (a) PAN. (b) LR MS. (c) GS \cite{GS}. (d) IHS \cite{IHS}. (e) Brovey \cite{Brovey}. (f) HPF \cite{HPFandHPFC}. (g) LMM \cite{LMVMandLMM}. (h) LMVM \cite{LMVMandLMM}. (i) SFIM \cite{SFIM}. (j) PNN \cite{PNN}. (k) DRPNN \cite{DRPNN}. (l) MSDCNN \cite{MSDCNN}. (m) PanNet \cite{PanNet}. (n) PSGAN \cite{PSGAN}. (o) Pan-GAN \cite{Pan-GAN}. (p) Ours. }
\label{fig::WV results}
\end{figure*}

\subsubsection{Quantitative Results}
We provide quantitative comparisons of the 14 methods on the two datasets. For the non-reference metrics, $i.e.$, D$_\lambda$, D$_s$, and QNR, the testing is on the full-scale images where the original MS and PAN images are used as the inputs. For the reference metrics, $i.e.$, SAM, ERGAS, and SSIM, the testing is under the Wald's Protocol \cite{wald} where the original MS and PAN images are downsampled before feeding into models and the original MS images serve as the ground truth.

From Table~\ref{tab::GF-2 result} we can see, our method achieves better results than Pan-GAN \cite{Pan-GAN} in terms of all metrics. When considering the traditional methods, the superiority is still significant. Except for D$_\lambda$, LMVM \cite{LMVMandLMM} reaches a bit better results (0.004) than ours (0.005). However, the supervised methods still keep their superiorities on reference metrics, because they can directly learn from the ground truth images while unsupervised methods cannot receive any guide information from the ground truth. Nevertheless, our method achieves the best performance on full-scale testing compared with all others, especially for the QNR metric, where our method reaches the best value of 0.982 with a significant gain of 0.019 when compared with the second best result. 

Table~\ref{tab::WV-3 result} displays the quantitative assessments on WorldView-3 dataset. Compared to GaoFen-2 images, WorldView-3 images are in higher spatial resolution and deeper bit depth, and the source images are less diverse as shown in Table~\ref{tab::dataset}, which makes the learning-based methods hard to train. We can see that results of all methods get worse, especially for those supervised methods. Among the unsupervised methods, the performance of our model on reference metrics becomes worse than Pan-GAN. However, we still keep the best results on all non-reference metrics compared with other methods, especially for the QNR metric, where our method reaches the best value of 0.933 with a significant gain of 0.023 when compared with the second best result. 

From the two tables, we can conclude that our proposed method achieves the best results on full-scale images and satisfactory results under Wald's evaluation protocol, which demonstrates its practical value in the real-world application.

\subsubsection{Visual Results}
Fig.~\ref{fig::GF results} and Fig.~\ref{fig::WV results} show some pan-sharpened examples. The results are produced under the full-scale setting. 

It can be observed that the traditional methods tend to distort colors and generate low quality spatial details. On GaoFen-2 dataset, Figs.~\ref{fig::gf_Brovey} and \ref{fig::gf_LMM} show that both Brovey and LMM suffer from spectral distortions. HPF and LMVM (Figs.~\ref{fig::gf_HPF} and \ref{fig::gf_LMVM}) cannot keep the spatial details well. On WorldView-3 dataset, GS (Fig.~\ref{fig::wv_GS}), IHS (Fig.~\ref{fig::wv_IHS}), and Brovey (Fig.~\ref{fig::wv_Brovey}) tend to distort spectral information. LMVM (Fig.~\ref{fig::wv_LMVM}) generates blurry results on WorldView-3 dataset.

From Fig.~\ref{fig::GF results} we can see that the supervised methods cannot transfer their knowledge learned from down-scaled setting very well to the real-world scenarios. DRPNN (Fig.~\ref{fig::gf_drpnn}), PanNet (Fig.~\ref{fig::gf_pannet}), and PSGAN (Fig.~\ref{fig::gf_psgan}) seem to generate blurry results and cannot preserve clear spatial details from the input PAN image (Fig.~\ref{fig::gf_pan}). Moreover, the results are worse on the WorldView-3 satellite because it has deeper bit depth and higher spatial resolution. Supervised methods tend to have significant color distortions and poor spatial details (see Figs.~\ref{fig::wv_pnn}, \ref{fig::wv_drpnn}, \ref{fig::wv_msdcnn}, \ref{fig::wv_pannet}, and \ref{fig::wv_psgan}). 
 
From Figs.~\ref{fig::GF results} and \ref{fig::WV results}, it can be concluded that unsupervised methods work well on the full-scale images due to the knowledge learned from the original data. Our method generates pan-sharpened images (Figs.~\ref{fig::wv_model90-2}) with better spatial and color quality, showing it has remarkable ability of generating satisfactory fusion results in the real-world scenarios.

\begin{figure*}[t]
\flushright
	\subfigure[]{\label{fig::qb_pan}
		\includegraphics[width=3.5cm]{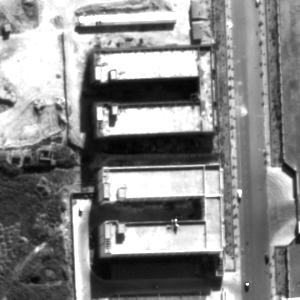}} \hspace{-9pt}
	\subfigure[]{\label{fig::qb_lrms}
		\includegraphics[width=3.5cm]{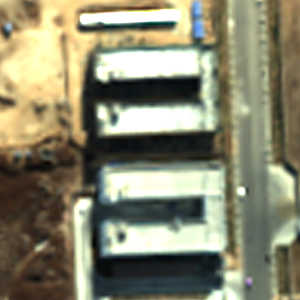}} \hspace{-9pt}
	\subfigure[]{\label{fig::qb_pnn}
		\includegraphics[width=3.5cm]{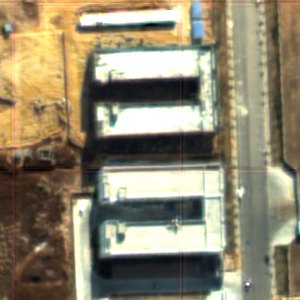}} \hspace{-9pt}
	\subfigure[]{\label{fig::qb_drpnn}
		\includegraphics[width=3.5cm]{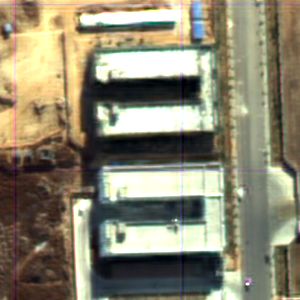}} \hspace{-9pt}
	\subfigure[]{\label{fig::qb_msdcnn}
		\includegraphics[width=3.5cm]{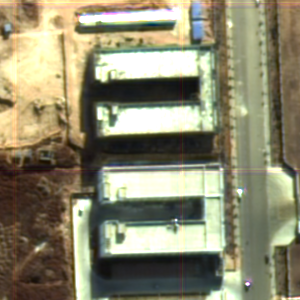}} \\
		\vspace{-8pt}
	\subfigure[]{\label{fig::qb_pannet}
		\includegraphics[width=3.5cm]{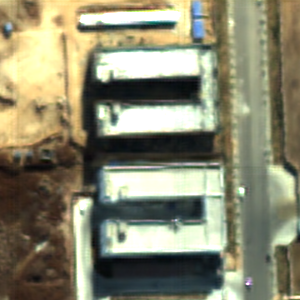}} \hspace{-9pt}
	\subfigure[]{\label{fig::qb_psgan}
		\includegraphics[width=3.5cm]{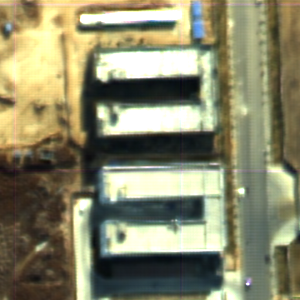}} \hspace{-9pt}
	\subfigure[]{\label{fig::qb_pan_gan}
		\includegraphics[width=3.5cm]{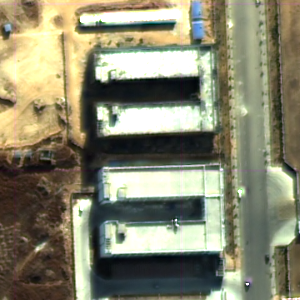}} \hspace{-9pt}
	\subfigure[]{\label{fig::qb_model90-2}
		\includegraphics[width=3.5cm]{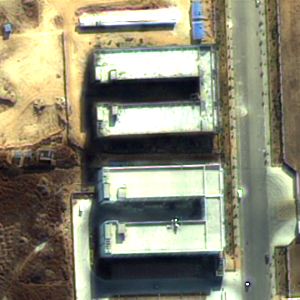}} \\	
		\vspace{-8pt}
\caption{Results of the generalization experiments. Models are trained on GaoFen-2 dataset and tested on QuickBird dataset. Visualized in RGB. (a) PAN. (b) LR MS. (c) PNN \cite{PNN}. (d) DRPNN \cite{DRPNN}. (e) MSDCNN \cite{MSDCNN}. (f) PanNet \cite{PanNet}. (g) PSGAN \cite{PSGAN}. (h) Pan-GAN \cite{Pan-GAN}. (i) Ours.}
\label{fig::QB results}
\end{figure*}

\subsection{Generalization Study}
To examine the robustness of deep learning models across different satellites, we conduct experiments about generalization ability. We generalize the learning based models that trained on GaoFen-2 dataset to a new satellite, QuickBird. 

Fig.~\ref{fig::QB results} shows some examples of the pan-sharpened results. When generalizing to a new satellite, PNN (\ref{fig::qb_pnn}), DRPNN (\ref{fig::qb_drpnn}), and MSDCNN (\ref{fig::qb_msdcnn}) suffer from notable spectral distortions, where we can observe some abnormal color lines. Besides, PanNet (\ref{fig::qb_pannet}) and PSGAN (\ref{fig::qb_psgan}) cannot generalize well to unsee data as well. Both of them generate blurry images with poor spatial details. On the other hand, unsupervised models are robust to different satellites. Compared to Pan-GAN (\ref{fig::qb_pan_gan}), the result of our method (\ref{fig::qb_model90-2}) preserve better spectral and spatial details. The reason may be that our network makes full use of the high-frequency content to enhance the spatial details in pan-sharpened results, thus it can generalize better to new satellites. Moreover, as an unsupervised model, we focus on reconstructing the pan-sharpened from input images and the consistency with the input images rather than the simulated ground truth in supervised methods. Therefore, our model can learn satellite-agnostic knowledge well.

\begin{table}[t]\vspace{-10pt}
\centering
  \caption{Study of all comparative methods w.r.t. inference time and the number of trainable parameters. Note that the pan-sharpened images are with a shape of $400\times400\times4$, we give an average time on them. As for the GAN-based models, we only give parameters of the generator.}
  \label{tab::efficiency study}
  \begin{tabular}{r|r||c|c}
    \toprule
    Processor & Method & Time(s) & $\sharp$Params \\
    \midrule
				& GS	 \cite{GS} 			& 0.1205 & - \\
    			& IHS \cite{IHS} 		& 0.0121 & - \\
    Intel Core  & Brovey \cite{Brovey}	& 0.0125 & - \\
    i7-7700HQ 	& HPF \cite{HPFandHPFC}	& 0.1061 & - \\
    CPU@2.80GHz & LMM \cite{LMVMandLMM}	& 0.1062 & - \\
    			& LMVM \cite{LMVMandLMM}	& 0.6800 & - \\
    			& SFIM \cite{SFIM}		& 0.1063 & - \\
    \midrule
    			& PNN \cite{PNN}			& 0.0002 & $\sim$ 0.080M \\
    			& DRPNN \cite{DRPNN}		& 0.0254 & $\sim$ 1.639M \\
    NVIDIA 		& MSDCNN \cite{MSDCNN}	& 0.0007 & $\sim$ 0.262M \\
    GeForce 	& PanNet \cite{PanNet} 	& 0.0006 & $\sim$ 0.077M \\
    RTX 2080Ti 	& PSGAN 	\cite{PSGAN} 	& 0.0009 & $\sim$ 1.654M \\
    			& Pan-GAN \cite{Pan-GAN}	& 0.0003 & $\sim$ 0.092M \\
				& Ours 					& 0.0016	 & $\sim$ 0.195M \\
	\bottomrule
\end{tabular}
\end{table}

\subsection{Efficiency Study}
To evaluate the efficiency of the pan-sharpening methods, we test the inference time of each method. All deep learning based methods are implemented in PyTorch \cite{PyTorch} framework, trained and tested on a single NVIDIA GeForce RTX 2080Ti GPU. As for the traditional methods, all of them are implemented using Matlab and run on an Intel Core i7-7700HQ CPU. To obtain the inference time of each model, we input the models with 286 test patch pairs and calculate the average running time of them. The efficiency results are illustrated in Table~\ref{tab::efficiency study}. Brovey \cite{Brovey} and IHS \cite{IHS} are the fastest traditional methods and they take about 0.012 seconds to generate one image. HPF \cite{HPFandHPFC}, LMM \cite{LMVMandLMM} and SFIM \cite{SFIM} cost around 0.1 seconds for one image. GS \cite{GS} needs 0.12 seconds. LMVM \cite{LMVMandLMM} is the slowest method among the traditional methods which cost 0.68 seconds to solve one pan-sharpened image. Thanks to the powerful parallel acceleration capability of GPU architectures, deep learning models run fast. PNN \cite{PNN} is the fastest models while PanNet \cite{PanNet} contains the least parameters. Nearly all deep learning models spend around 0.001 seconds for generating one fusion image, except DRPNN \cite{DRPNN} who is much slower and cost around 0.025 seconds due to its deeper network architecture and bigger size of filters. Our proposed model has an acceptable time consuming and memory cost compared with other deep learning based models. It is efficient to solve the pan-sharpening task. 

\section{Conclusion} \label{section::Discussion}
In this work, we have presented a novel unsupervised generative adversarial model for pan-sharpening, termed as UCGAN. We explored several architectures and designs to find the optimal solution for pan-sharpening in an unsupervised manner. Under the unsupervised framework, our proposed method can be trained on full-scale images without ground truth and make full use of the original information of the raw data. For keeping the consistency with the real application scenarios, we focus on original PAN and LR MS images without any preprocessing step. Furthermore, our architecture and loss function improve the quality of the fusion result, especially on the full-scale images. Ablation studies and comparison with other state-of-art methods demonstrate the superiority of our model in the real-world application. 

Despite the great improvement on the full-scale images, there is still a gap to supervised models. The supervision information in down-scaled images is an inherent advantage for the supervised models. To minimize this gap, we will go deeply into the potentiality of the unsupervised method for pan-sharpening and optimize its architecture design and improve its performance in our future work.



\ifCLASSOPTIONcaptionsoff
  \newpage
\fi



%
\bibliographystyle{IEEEtran}
\bibliography{main}

%








\end{document}